\documentclass[letterpaper,10pt]{article}
\usepackage[margin=20truemm]{geometry}

\usepackage{microtype}
\usepackage{graphicx}
\usepackage{booktabs} 
\usepackage{multirow}
\usepackage{multicol}
\usepackage{array}
\usepackage{threeparttable}
\usepackage{natbib}

\usepackage{hyperref}
\usepackage{here}

\usepackage{amsmath}
\usepackage{amssymb}
\usepackage{mathtools}
\mathtoolsset{showonlyrefs=true} 
\usepackage{amsthm}
\usepackage[whole]{bxcjkjatype} 

\usepackage[inline]{enumitem}
\usepackage{listings}
\lstdefinestyle{chattemplate}{
  basicstyle=\ttfamily\tiny,
  columns=fullflexible,
  breaklines=true,
  frame=single,
  keepspaces=true,
  showstringspaces=false
}
\hypersetup{
  colorlinks   = true, 
  urlcolor     = [rgb]{0,0,1}, 
  linkcolor    = [rgb]{0,0,0.5}, 
  citecolor   = [rgb]{0,0,0.5} 
}

\makeatletter
\let\MYcaption\@makecaption
\makeatother

\usepackage{subcaption}
\makeatletter
\let\@makecaption\MYcaption
\makeatother

\usepackage[capitalize,noabbrev]{cleveref}
\usepackage{relsize}

\theoremstyle{plain}
\newtheorem{theorem}{Theorem}
\newtheorem{proposition}[theorem]{Proposition}

\theoremstyle{definition}

\theoremstyle{remark}

\crefname{theorem}{Theorem}{Theorems}
\Crefname{theorem}{Theorem}{Theorems}

\crefname{proposition}{Proposition}{Propositions}
\Crefname{proposition}{Proposition}{Propositions}

\crefname{lemma}{Lemma}{Lemmas}
\Crefname{lemma}{Lemma}{Lemmas}

\crefname{corollary}{Corollary}{Corollaries}
\Crefname{corollary}{Corollary}{Corollaries}

\crefname{definition}{Definition}{Definitions}
\Crefname{definition}{Definition}{Definitions}

\crefname{assumption}{Assumption}{Assumptions}
\Crefname{assumption}{Assumption}{Assumptions}

\crefname{remark}{Remark}{Remarks}
\Crefname{remark}{Remark}{Remarks}

\usepackage[textsize=tiny]{todonotes}

\newcommand{\argmax}{\mathop{\rm arg~max}\limits}

\usepackage{authblk}
\author[1]{Akiyoshi Tomihari\thanks{tomihari@g.ecc.u-tokyo.ac.jp}}
\affil[1]{Department of Computer Science, The University of Tokyo}
\date{}

\title{Learning Dynamics in RL Post-Training for Language Models}

\begin{document}
\maketitle
    \allowdisplaybreaks

\begin{abstract}
Reinforcement learning (RL) post-training is a critical stage in modern language model development, playing a key role in improving alignment and reasoning ability.
However, several phenomena remain poorly understood, including the reduction in output diversity.
To gain a broader understanding of RL post-training, we analyze the learning dynamics of RL post-training from a perspective that has been studied in supervised learning but remains underexplored in RL.
We adopt an empirical neural tangent kernel (NTK) framework and decompose the NTK into two components to characterize how RL updates propagate across training samples.
Our analysis reveals that limited variability in feature representations can cause RL updates to systematically increase model confidence, providing an explanation for the commonly observed reduction in output diversity after RL post-training.
Furthermore, we show that effective learning in this regime depends on rapidly shaping the classifier, which directly affects the gradient component of the NTK.
Motivated by these insights, we propose classifier-first reinforcement learning (CF-RL), a simple two-stage training strategy that prioritizes classifier updates before standard RL optimization.
Experimental results validate our theoretical analysis by demonstrating increased model confidence and accelerated optimization under CF-RL.
Additional analysis shows that the mechanism underlying CF-RL differs from that of linear-probing-then-fine-tuning in supervised learning.
Overall, our study formalizes the learning dynamics of RL post-training and motivates further analysis and improvement.

\end{abstract}

\section{Introduction}
Post-training aims to refine and adapt pretrained language models to specific tasks or user requirements, and reinforcement learning (RL) plays a central role in this phase.
In particular, reinforcement learning from human feedback (RLHF) has been shown to substantially improve instruction following, safety, and overall usefulness of modern language models~\citep{christiano2017deep,ziegler2019fine}.
More recently, RL post-training has also demonstrated strong gains in reasoning performance, often under the framework of reinforcement learning with verifiable rewards (RLVR)~\citep{guo2025deepseek,jaech2024openai,team2025kimi}.

However, many aspects of RL post-training remain poorly understood.
Prior work has shown that RLHF tends to reduce output diversity, leading to more concentrated output distributions~\citep{kirkunderstanding,chung2025modifying}.
Similarly, recent studies indicate that RLVR may not elicit fundamentally new reasoning patterns~\citep{yue2025does}.
These findings highlight significant gaps in our understanding of how RL post-training reshapes model behavior.

In this paper, we address these gaps by analyzing the learning dynamics of RL post-training.
Learning dynamics characterizes how updates driven by individual training examples influence the model's predictions on other examples, a perspective that has been extensively studied in supervised learning~\citep{ren2025learning,ren2022better,ren2023how,chen2023which}, but remains underexplored in RL.
To this end, we adopt an empirical neural tangent kernel (NTK) framework and decompose the NTK into two components~\citep{tomihari2024understanding}.
Our analysis reveals that RL updates can systematically increase model confidence, which in turn may explain the observed reduction in output diversity through probability concentration.
Furthermore, our analysis highlights the importance of rapidly shaping the gradient component, which explicitly involves the classifier.
This observation motivates a classifier-first optimization strategy, inspired by linear-probing-then-fine-tuning (LP-FT) in supervised fine-tuning~\citep{kumar2022finetuning}, where the classifier is optimized prior to full model updates.

Our contributions are summarized as follows:
\begin{itemize}
    \item We formalize the learning dynamics of RL post-training through a decomposition of the empirical neural tangent kernel (NTK) (\cref{sec:training}, \cref{prop:dynamics}).
    \item Based on this formalization, we explain how RL post-training can lead to increased model confidence due to high similarity in feature representations (\cref{sec:failure}, \cref{prop:representation}).
    \item Motivated by this analysis, we highlight the importance of rapidly shaping the classifier and propose a classifier-first reinforcement learning (CF-RL).
\end{itemize}

\section{Related work}
\paragraph{NTK and learning dynamics.}
The learning dynamics of neural networks are studied from various perspectives.
\citet{rahaman2019spectral} showed that neural networks tend to learn low-frequency components of target functions earlier than high-frequency ones. At the architectural level, \citet{chen2023which} observed a layer-wise convergence pattern, where shallower layers converge faster and capture coarse structure. \citet{park2024emergence} analyzed the learning dynamics of generative models in a conceptual space.

The neural tangent kernel (NTK), introduced by \citet{jacot2018neural}, provides a theoretical framework for analyzing the learning dynamics of infinitely wide neural networks. In the infinite-width limit, the NTK converges to a deterministic kernel that remains constant throughout training. Subsequent studies further characterized training dynamics in this regime~\citep{lee2019wide,arora2019exact}. Later studies showed that NTK-based analyses can be used to explain the behavior of real-world, finite-width models~\citep{wei2022more,ren2022better,mohamadi2022making}, in which the kernel depends on the network parameters and may evolve during training. This parameter-dependent kernel is commonly referred to as the empirical NTK~\citep{loo2022evolution,mohamadi2023fast}.

Building on this line of work, several studies have applied the empirical NTK to understand fine-tuning dynamics. \citet{malladi2023kernel} showed, both theoretically and empirically, that prompt-based fine-tuning exhibits behavior consistent with kernel predictions. \citet{ren2023how} and \citet{tomihari2024understanding} analyzed linear probing followed by fine-tuning (LP-FT)~\citep{kumar2022finetuning} through the lens of the empirical NTK. \citet{ren2025learning} provided a framework for supervised fine-tuning, highlighting a squeezing effect in off-policy DPO and the advantages of on-policy variants. Our work contributes to analyzing the learning dynamics of RL from the NTK perspective.

\paragraph{Classifier--feature interaction.}
The relationship between feature representations and classifiers (which we formalize in~\cref{sec:pre}) has been extensively studied. \citet{papyan2020prevalence} identified the phenomenon of ``neural collapse'' in classification tasks, where features converge to class means that are equinorm, equiangular, and aligned with the classifier, leading to improved generalization and robustness. \citet{wu2024linguistic} extended this phenomenon, showing that neural collapse-like behavior also emerges in language modeling, even when classical assumptions are violated. 

Motivated by neural collapse, \citet{yang2022inducing} proposed the equiangular tight frame classifier~\citep{strohmer2003grassmannian} to explicitly encourage neural collapse. In fine-tuning, \citet{kumar2022finetuning} analyzed why directly fine-tuning well-trained representations with a randomly initialized classifier can degrade out-of-distribution generalization, and proposed LP-FT as a remedy. For language models, \citet{tomihari2024understanding} highlighted the importance of classifier norms in shaping fine-tuning dynamics.
\citet{razinunintentional} and \citet{razin2025your} analyzed likelihood displacement phenomena in DPO and implicit reward models, respectively, both emphasizing the role of the classifier.
In contrast to these works, we study RL post-training with a particular focus on how the classifier shapes RL updates.

\paragraph{RL post-training.}
RL post-training is a central paradigm for aligning large language models with human preferences, most prominently through reinforcement learning from human feedback (RLHF)~\citep{ziegler2019fine,ouyang2022training}. RLHF has played a key role in producing models that are safer and more helpful, and is now deployed in real-world systems~\citep{achiam2023gpt,openai_chatgpt,anthropic_claude}. It enables alignment with objectives that are difficult to specify using supervised data or hand-crafted reward functions. Beyond alignment, RL post-training has also been applied to enhance reasoning capabilities, often under the framework of RLVR~\citep{guo2025deepseek,jaech2024openai,team2025kimi}.

Despite its empirical success, several studies have identified limitations of RL post-training. \citet{kirkunderstanding} showed that while RLHF improves out-of-distribution generalization, it can substantially reduce output diversity. To mitigate this issue, \citet{chung2025modifying} proposed deviation-weighted preference optimization. In the context of RL with verifiable rewards (RLVR), \citet{yue2025does} showed that RL primarily reweights existing solutions rather than expanding the model’s underlying reasoning capacity. This observation is consistent with the findings of \citet{zhao2025echo}, who showed that RL algorithms tend to amplify patterns already present in pretraining.

From a theoretical perspective, \citet{wang2023rlhf} showed that preference-based RL can be reduced to reward-based RL, despite preference signals containing less information than explicit rewards.
Several studies have analyzed specific preference-based RL algorithms proposed in their studies, including dueling bandit-based~\citep{xu2020preference} and Bayesian approaches~\citep{novoseller2020dueling}.
Complementing these theory-driven studies, \citet{razin2024vanishing} and \citet{razin2025what} demonstrated both experimentally and theoretically that policy gradients can vanish when reward variance is low, and highlighted fundamental limitations of evaluating reward models solely by accuracy.
In this work, we analyze the learning dynamics of RL post-training from both experimental and theoretical perspectives.


\section{Preliminaries}
\label{sec:pre}
\subsection{Problem setup and notation}

\paragraph{Notation and setup.}
For a matrix $A$, we write $A_{i,j}$ for its $(i,j)$-th entry, and
$A_{i,:}$ and $A_{:,j}$ for its $i$-th row and $j$-th column, respectively.
For a vector $a$, $a_i$ denotes its $i$-th element.
To simplify notation, we sometimes use brackets $[\cdot]$ to denote indexing of vectors,
e.g., $[\pi_\theta(\cdot \mid x,y_{<l})]_{y_l}$. 

Let $\mathcal{X}$ denote the space of input prompts and $\mathcal{Y}$ the space of generated responses.
Both spaces consist of sequences of numerical token representations, such that
$\mathcal{X}, \mathcal{Y} \subseteq \{1,2,\ldots,V\}^{*}$,
where $V$ is the vocabulary size.

We consider RL post-training applied to a pre-trained language model
$\pi_\theta$ with parameters $\theta$, which has been fine-tuned via SFT
prior to RL.

\paragraph{Autoregressive generation.}
Given a prompt $x\in\mathcal{X}$, the model generates a response $y=(y_1,\ldots,y_{|y|})$
autoregressively.
At each position $l$, the conditional distribution over the next token is
\[
\pi_\theta(\cdot\mid x,y_{<l}) \in \Delta^{V-1},
\]
where $y_{<l}$ denotes the prefix of length $l-1$, and
\[
\Delta^{V-1} \coloneqq \{p\in\mathbb{R}^V_{\ge 0} \mid \sum_{v=1}^V p_v = 1\}
\]
denotes the $(V-1)$-dimensional probability simplex.
We write
\[
\pi_\theta(y\mid x) := \prod_{l=1}^{|y|} [\pi_\theta(\cdot\mid x,y_{<l})]_{y_l}.
\]

\paragraph{Model architecture.}
The input sequence $(x,y_{<l})$ is mapped to a $D$-dimensional feature representation
$\phi(x,y_{<l})\in\mathbb{R}^D$ (e.g., by a Transformer).
A linear classifier (unembedding matrix) $W\in\mathbb{R}^{V\times D}$ produces the logits,
and the predictive distribution is given by
\[
\pi_\theta(\cdot\mid x,y_{<l}) = \mathrm{softmax}\bigl(W\phi(x,y_{<l})\bigr).
\]

\subsection{RL post-training}

Let $r:\mathcal{X}\times\mathcal{Y}\to[-1,1]$ denote a reward model
that assigns a scalar score to each input--output pair $(x,y)$.
In RL post-training, the goal is to maximize the expected reward while constraining
the model to remain close to a reference model $\pi_{\theta_{\mathrm{ref}}}$, which corresponds to the model immediately after SFT.

The objective is defined as
\begin{align}
J(\theta)
&:= \frac{1}{N}\sum_{i=1}^N
\mathbb{E}_{y\sim\pi_\theta(\cdot\mid x_{i})}[\hat r(x_{i},y)],\\
\hat r(x,y)
&:= r(x,y) - \lambda\,\mathrm{KL}\!\left(
\pi_\theta(\cdot\mid x)\,\|\,\pi_{\theta_{\mathrm{ref}}}(\cdot\mid x)
\right),
\end{align}
where $\lambda\geq 0$ controls the strength of KL regularization.

We use the subscript $t$ to indicate quantities evaluated at training
step $t$. The model parameters are updated via gradient ascent,
\begin{align}
\theta_{t+1} = \theta_t + \eta \nabla_\theta J(\theta_t),
\end{align}
where $\eta$ is the learning rate, which is typically small
\citep{ouyang2022training}.

In practice, the expectation and gradient are approximated using
policy optimization algorithms such as PPO~\citep{schulman2017proximal},
GRPO~\citep{shao2024deepseekmath}, or
RLOO~\citep{kool2019buy,ahmadian2024back}.

\section{Learning dynamics of RL post-training}
This section analyzes RL post-training dynamics using the empirical NTK.
We characterize stepwise output changes, identify limitations of representation-based updates, and motivate classifier-first reinforcement learning (CF-RL).

\subsection{Stepwise change of the model output}
\label{sec:training}
Following the fine-tuning analysis of \citet{ren2025learning}, we characterize the stepwise evolution of the model output under reward-based optimization.
We focus on how the log-probability of the model output changes across training steps.

For a given prompt $x \in \mathcal{X}$ and a self-generated prefix $y_{<m} \in \mathcal{Y}$ of length $m-1$, let $\chi_{<m} \coloneqq (x, y_{<m})$. 

We define
{\relsize{-0.5}
\begin{align}
\Delta_t \log \pi(\cdot \mid \chi_{<m})
\coloneqq
\log \pi_{\theta_{t+1}}(\cdot \mid \chi_{<m})
-
\log \pi_{\theta_t}(\cdot \mid \chi_{<m}),
\end{align}}\noindent
which measures the change in the model's output distribution for the $m$-th token between training steps.

To simplify notation, we introduce
\begin{align}
    T_{x,y_{<m}}
    &\coloneqq
    I_{V} - \mathbf{1}\,\pi_{\theta_t}(\cdot \mid x,y_{<m})^{\!\top}, \quad 
    d_{y_{i},l}
    \coloneqq
    e_{y_{i,l}} - \pi_{\theta_t}(\cdot \mid x_{i}, y_{i,<l}).
\end{align}

\begin{proposition}
\label{prop:dynamics}
The change in the model output can be written as
\begin{align}
    \Delta_{t}\log \pi(\cdot \mid x, y_{<m})
    =  \frac{\eta}{N}\sum_{i=1}^{N}
       \mathbb{E}_{y_{i} \sim \pi_{\theta_{t}}(\cdot \mid x_{i})}
       \Biggl[
           \sum_{l=1}^{|y_{i}|}
           \hat{r}(x_{i},y_{i})\,
           T_{x, y_{<m}}\,\mathcal{K}_{t}(x,y_{<m},x_{i},y_{i,<l})\,d_{y_{i},l}
       \Biggr]
       + O\!\left(\eta^{2}\bigl\|\tfrac{\partial J(\theta_{t})}{\partial \theta}\bigr\|^{2}\right),\label{eq:dynamics}
\end{align}
Here $\mathcal{K}_{t}(x,y_{<m},x_{i},y_{i,<l})$ is the empirical NTK which is decomposed as 
\begin{align}
     R_{t}(x,y_{<m},x_{i},y_{i,<l}) + G_{t}(x,y_{<m},x_{i},y_{i,<l}),
\end{align}
where $R_{t}(x,y_{<m},x_{i},y_{i,<l})$ is the Representation component
\begin{align}
    \langle\phi_{t}(x,y_{<m}), \phi_{t}(x_{i}, y_{i,<l})\rangle I_{V}
\end{align}
and $G_{t}(x,y_{<m},x_{i},y_{i,<l})$ is the Gradient component 
\begin{align}
    W_{t}\,\frac{\partial \phi_{t}(x, y_{<m})}{\partial\theta^{\phi}}\left(\frac{\partial \phi_{t}(x_{i},y_{i,<l})}{\partial\theta^{\phi}}\right)^{\!\top} W_{t}^{\top}.
\end{align}
\end{proposition}
\paragraph{Interpretation of the update.}
This proposition expresses the update of the model output at $(x, y_{<m})$ as a weighted aggregation of contributions from individual training samples $(x_{i}, y_{i})$.
The empirical NTK $\mathcal{K}_t$ therefore acts as a model-induced similarity measure between samples, consistent with the interpretation of \citet{ren2025learning}.

\paragraph{Higher-order terms.}
The remainder term
$O(\eta^{2}\bigl\|\tfrac{\partial J(\theta_{t})}{\partial \theta}\bigr\|^{2})$ is negligible in practice because RLHF typically uses a small learning rate~\citep{ouyang2022training,ahmadian2024back,wen2025rethinking}. In addition, the gradient norm $\|\partial J(\theta_{t})/ \partial \theta\|$ is often kept small by standard optimization techniques such as gradient clipping, and analyses sometimes consider the limit $\eta \to 0$~\citep{razin2025what}.  

\paragraph{NTK decomposition.}
The decomposition of the empirical NTK into the Representation and Gradient components follows \citet{tomihari2024understanding}. The Representation component measures similarity via the inner product of feature vectors, whereas the Gradient component captures similarity through the gradients of the feature map $\phi_{t}(\cdot)$. In the following, we analyze how these two components contribute differently to the learning dynamics.

\subsection{Failure of the Representation component}
\label{sec:failure}
To gain insight into the behavior of the Representation component, we examine empirical statistics of the feature vectors $\phi_t(\cdot)$.
Table~\ref{tab:feature} summarizes these statistics, with additional details provided in Appendix~\ref{app:detail_fig}.

We observe that the cosine similarity between feature vectors is often high.
This suggests that feature vectors exhibit directional alignment, rather than being uniformly distributed in $\mathbb{R}^D$.
This behavior is reminiscent of rank-collapse phenomena reported in Transformer architectures~\citep{dong2021attention,noci2022signal}.

\begin{table}[tbp]
    \centering
    \caption{
        Distribution of feature cosine similarities across samples. We report the mean, standard deviation, and minimum cosine similarity between feature representations computed from distinct input–output samples.
    }
    \label{tab:feature}
    \begin{tabular}{lccc}
        \toprule
        Model & Mean & Std & Min \\
        \midrule
        Pythia-2.8B   & $0.606$ & $0.085$ & $0.247$ \\
        Qwen2.5-3B &  $0.656$ & $0.324$ &  $ 0.042$\\
        \bottomrule
    \end{tabular}
\end{table}

Here, we will see that this strong feature alignment limits the discriminative power of the resulting update.
To isolate the contributions of the Representation and Gradient component, we decompose the update into the following two components.
{\relsize{-0.5}
\begin{align}
    u^{\mathrm{Rep}}_{t,l}(x,y_{<m},x_{i},y_{i})
    &\coloneqq
    T_{x, y_{<m}} R_{t}(x,y_{<m},x_{i},y_{i,<l}) d_{y_{i},l},
    \\
    u^{\mathrm{Grad}}_{t,l}(x,y_{<m},x_{i},y_{i})
    &\coloneqq
    T_{x, y_{<m}} G_{t}(x,y_{<m},x_{i},y_{i,<l}) d_{y_{i},l}.
\end{align}}
Using these definitions, the quantity inside the expectation in the first
term of Eq.~\eqref{eq:dynamics},
\[
    \sum_{l=1}^{|y_{i}|}
        \hat{r}(x_{i}, y_{i})\,
        T_{x, y_{<m}}\,\mathcal{K}_{t}(x, y_{<m}, x_{i}, y_{i,<l})\,d_{y_{i},l},
\]
namely the reward-weighted update contribution of sample $(x_{i}, y_{i})$ aggregated
over all token positions, can be rewritten as
\[
    \sum_{l=1}^{|y_{i}|}
        \hat{r}(x_{i}, y_{i})\,
        (u^{\mathrm{Rep}}_{t,l}(x,y_{<m},x_{i},y_{i}) + u^{\mathrm{Grad}}_{t,l}(x,y_{<m},x_{i},y_{i})).
\]
The following proposition highlights the limited discriminative power of the Representation component, motivated by the empirical observations in the previous subsection.
\begin{proposition}
\label{prop:representation}
    Assume that the feature inner product satisfies
    \begin{align}
        \langle\phi_{t}(x,y),\phi_{t}(x',y')\rangle &\geq 0
    \end{align}
    for any $x, y, x', y' \in \mathcal{V}$.
    Then the update induced by the Representation component satisfies
    \begin{align}
        \argmax_{1 \le v \le V} [u^{\mathrm{Rep}}_{t,l}(x,y_{<m},x_{i},y_{i})]_{v} = y_{i,l}.
    \end{align}
\end{proposition}
This result indicates that the Representation component reinforces the reward signal by increasing the predicted probability of the sampled token $y_{i,l}$.
Specifically, the maximal entry of $u^{\mathrm{Rep}}_{t,l}(x,y_{<m},x_{i},y_{i})$ always corresponds to $y_{i,l}$, independent of the conditioning context $(x, y_{<m})$.
This assumption is mild for models exhibiting strongly aligned feature representations, as suggested by the empirical cosine similarity statistics in~\Cref{tab:feature}.

\paragraph{Increased model confidence.}
When $\hat{r}(x_i, y_i) > 0$, since $y_i$ is drawn from the model distribution, i.e., $y_i \sim \pi_{\theta_t}$, the resulting update amplifies the model's existing preference.
As a consequence, the Representation component tends to increase the model's confidence in its sampled outputs by further sharpening the output distribution.

\begin{figure}[tp]
    \centering
    \includegraphics[width=0.9\linewidth]{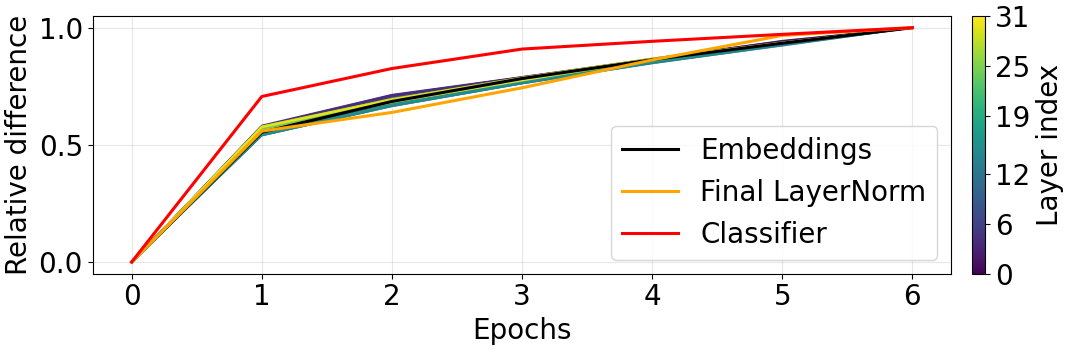}
    \caption{
\textbf{The classifier learns faster than the other parameters.}
We partition the model parameters into transformer layers (layers $0$--$31$), token embeddings (Embeddings), the final layer normalization (Final LayerNorm), and the classifier.
For each group, we plot the L2 norm of the parameter difference from the SFT model during RL training, scaled so that the norm equals $1$ at the end of RL.
}
\label{fig:weight_change}
\end{figure}

\subsection{The Gradient component and classifier-driven learning}
\label{sec:CF-RL}

The preceding analysis shows that when feature similarities are high, the Representation component provides only a limited notion of similarity and induces highly constrained update directions.
Since effective learning requires the empirical NTK to capture richer, sample-dependent similarity structures, this limitation necessitates reliance on the Gradient component.

\paragraph{Expressivity of the Gradient component.}
\citet{tomihari2024understanding} shows that the Gradient component (referred to as the FT-effective component in their work) possesses greater expressive capacity than the Representation component.
Unlike the Representation component, the Gradient component incorporates the classifier matrix $W$ in addition to the gradient of the feature map $\phi$.
Notably, the classifier $W$ does not appear in the Representation component, underscoring its central role in shaping Gradient-based updates.

\paragraph{Empirical evidence: rapid learning of the classifier.}
If the Gradient component governs effective learning, parameters that directly shape this component should be optimized early.
To examine this, in \Cref{fig:weight_change}, we plot the relative parameter change
$\|\theta^{(g)}_{t} - \theta^{(g)}_{0}\|/\|\theta^{(g)}_{T} - \theta^{(g)}_{0}\|$
across training epochs for different parameter groups, including transformer layers, token embeddings, layer normalization, and the classifier.
We observe that the classifier parameters consistently exhibit the fastest relative change, indicating that they are learned earlier than other components.
Within our framework, this behavior is expected: learning an informative Gradient component requires rapid adaptation of the classifier to capture meaningful sample similarity.

\paragraph{Classifier-first reinforcement learning (CF-RL).}
Motivated by both the role of the classifier in the Gradient component and its empirically fast optimization, we hypothesize that prioritizing classifier updates can improve RL post-training.
This idea parallels the ``linear probing then fine-tuning'' (LP-FT)~\citep{kumar2022finetuning} strategy used in supervised fine-tuning, where the classifier is optimized before jointly training all model parameters.

Adapting this principle to reinforcement learning, we propose a two-stage training scheme termed
\emph{classifier-first reinforcement learning (CF-RL)}.
In the first stage, reinforcement learning is performed while freezing all parameters except the classifier.
In the second stage, standard RL training is resumed with all parameters updated jointly.

We posit that optimizing the classifier prior to full-scale RL yields a more informative Gradient-based kernel early in training, thereby improving the efficiency of subsequent RL optimization.

\begin{figure}[tp]
    \centering
    \begin{minipage}{0.65\linewidth}
    \centering
    \includegraphics[width=0.9\linewidth]{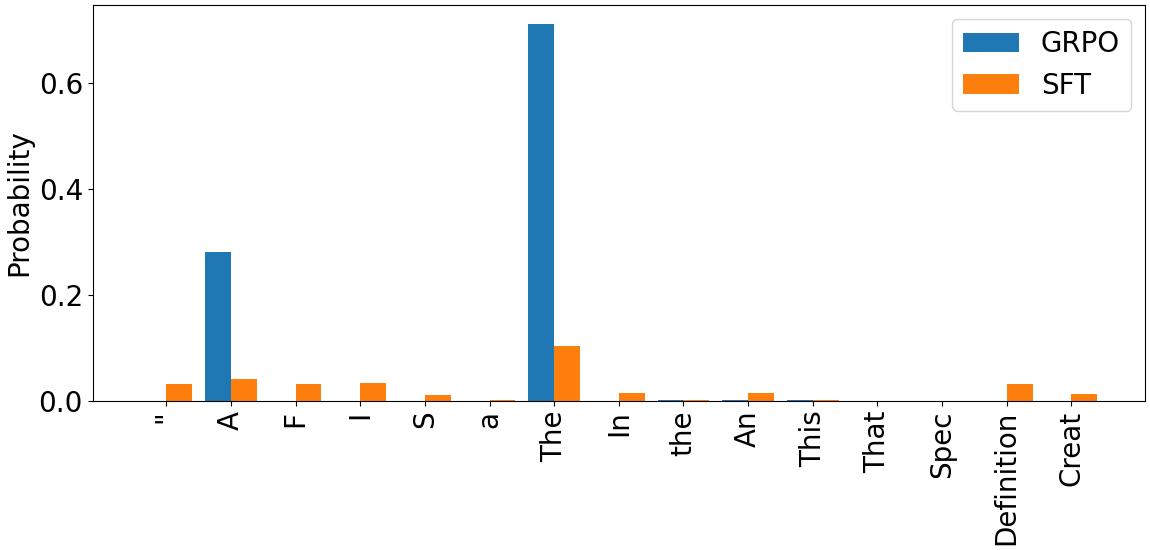}
    \subcaption{Example first-token distribution (SFT vs.\ GRPO)}
        \label{fig:first_token_example}
    \end{minipage}%
    \begin{minipage}{0.34\linewidth}
    \centering
    \includegraphics[width=0.96\linewidth]{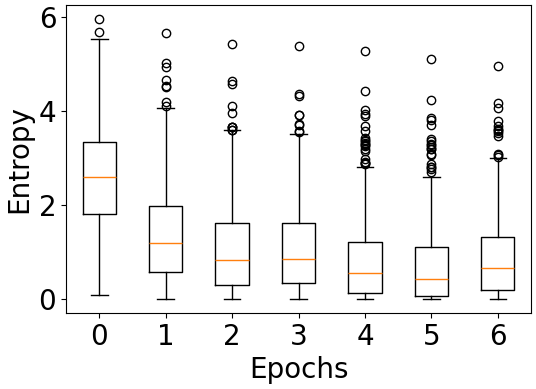}
    \subcaption{Entropy reduction during RL. Epoch~0 corresponds to the SFT model. Lower entropy indicates a more concentrated distribution.}
        \label{fig:first_token_entropy}
    \end{minipage}
    \caption{
\textbf{RL increases model confidence.}
(a): An example of the probability distribution of the first generated token for the prompt
\textsf{``Describe the given scene in a few words.\textbackslash n Two children playing with a ball on the beach.''}
The figure shows the union of the top-10 tokens by probability from each trained model, yielding 15 distinct tokens in total.
Compared to SFT, the distribution after GRPO is more concentrated.
(b): Boxplots of the entropy of the first-token distributions across prompts, where each point corresponds to a single prompt.
The entropy decreases monotonically during RLHF, indicating increasingly peaked output distributions.
}
\label{fig:confidence}
\end{figure}

\section{Experiments}
We experimentally validate our findings and conduct further analysis by examining:
{ \setlength{\leftmargini}{6pt} \setlength{\itemindent}{0pt}
\begin{itemize}
    \item Whether RL increases model confidence (\cref{sec:confidence}).
    \item Whether CF-RL accelerates RL optimization (\cref{sec:cf-rl}).
    \item How CF-RL influences the learning process (\cref{sec:cf-rl-analysis}).
\end{itemize}}

\subsection{Setting}

\paragraph{Overall protocol.}
Our experimental setup largely followed the RL post-training pipeline of \citet{razin2025what}.
Since our focus is on the optimization dynamics during the RL stage rather than reward modeling,
we do not train a reward model from human preference data.
Instead, we used a ground-truth reward model to provide supervision, as commonly done in prior work.
Specifically, we adopted ArmoRM~\citep{armorm} as a ground-truth reward model,
which produces scalar rewards and allows us to study RL post-training with general reward signals.

\paragraph{Base language model.}
For the policy model, we use the pre-trained Pythia 2.8B model~\citep{biderman2023pythia}.  
Following \citet{razin2025what}, we first perform SFT on AlpacaFarm~\citep{alpacafarm}, and then run RL training on UltraFeedback~\citep{cui2024ultrafeedback}. We used GRPO~\citep{shao2024deepseekmath} for the policy gradient algorithm. 

\begin{figure}
    \centering
    \begin{minipage}{0.48\linewidth}
        \includegraphics[width=0.9\linewidth]{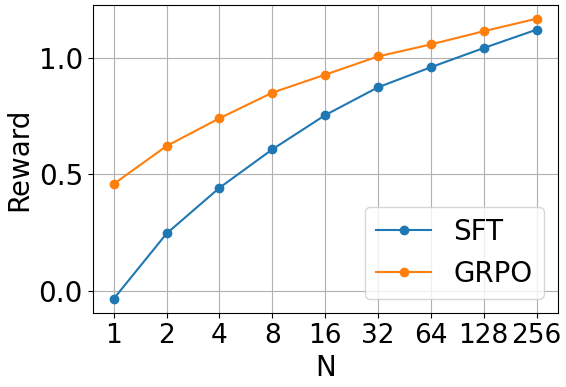}
        \subcaption{Best-of-$N$ rewards for SFT and GRPO.}
        \label{fig:bestofn}
    \end{minipage}
    \hfill
    \begin{minipage}{0.48\linewidth}
        \includegraphics[width=0.9\linewidth]{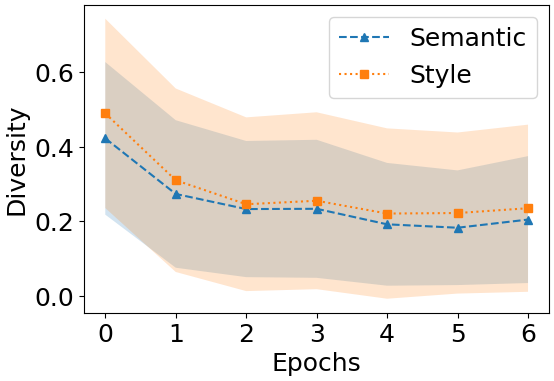}
        \subcaption{Semantic and style diversity over RL epochs. Higher values indicate greater diversity. Shaded regions denote standard deviation.}
        \label{fig:diversity}
    \end{minipage}
    \caption{\textbf{RL improves reward by concentrating probability mass, at the cost of reduced output diversity.}}
    \label{fig:BoN-diversity}
\end{figure}

\begin{table}[t]
\centering
\caption{\textbf{Lower entropy for high-reward samples.}
We split the training dataset based on reward values and performed one epoch of RL.}
\label{tab:first_token_entropy_reward}
\begin{tabular}{lc}
\toprule
\textbf{Data group} & \textbf{Entropy} \\
\midrule
Low-reward data  & $2.439 \pm 1.278$ \\
High-reward data & $1.516 \pm 0.931$ \\
\bottomrule
\end{tabular}
\end{table}

\subsection{RL increases model confidence}
\label{sec:confidence}
\paragraph{Entropy reduction of the output distribution.}
\Cref{fig:confidence} analyzes the probability distribution of the first generated token, together with representative examples and entropy measurements.
Because this distribution is conditioned solely on the input prompt, it is invariant across decoding trajectories and independent of subsequently generated tokens, enabling a fair comparison across training algorithms.
We observe that the distribution becomes increasingly concentrated after RL, which is quantitatively reflected by a monotonic decrease in entropy.
This indicates that RL systematically increases the model's confidence in its predictions.

\paragraph{Role of high-reward samples.}
To verify that the entropy reduction is driven by high-reward samples, we split the training data into two groups based on reward magnitude and perform one epoch of RL.
As shown in \Cref{tab:first_token_entropy_reward}, the high-reward group exhibits substantially lower first-token entropy than the low-reward group.
This result suggests that the overall reduction in entropy is primarily attributable to increased confidence on high-reward samples, rather than a uniform sharpening across all data.

\paragraph{Connection to reward improvement and diversity.}
To assess whether increased confidence contributes to reward gains, we compute the Best-of-$N$ metric in \Cref{fig:bestofn}.
The performance gap between SFT and GRPO narrows as $N$ increases, indicating that probability concentration partially explains the observed improvement in reward.
We further examine probability concentration through output diversity.
Following \citet{chung2025modifying}, we sample multiple outputs, embed them into semantic and style spaces, and compute pairwise cosine distances.
As shown in \Cref{fig:diversity}, diversity consistently decreases during RL training, confirming that the model outputs become increasingly similar.

\subsection{CF-RL accelerates RL optimization}
\label{sec:cf-rl}
\Cref{fig:cf-grpo} compares rewards across training epochs for CF-RL and standard RL.
Training only the classifier, corresponding to the first stage of CF-RL (epoch $0$), yields little reward improvement.
In contrast, the subsequent RL stage leads to a substantial reward gain, with the largest increase occurring at epoch $1$, consistent with our analysis of the Representation component in \Cref{sec:failure}.

\subsection{The learning process of CF-RL}
\label{sec:cf-rl-analysis}
We empirically investigate why CF-RL accelerates RL optimization.
In SFT, the effectiveness of the linear-probing-then-fine-tuning (LP-FT) strategy has been attributed to a reduction in feature distortion~\citep{kumar2022finetuning}, where an initial classifier-only stage suppresses unnecessary changes in feature representations.
Subsequent analyses further show that this reduction is closely associated with a substantial increase in classifier norms~\citep{tomihari2024understanding}.

\begin{figure}
\centering
\begin{minipage}{0.49\columnwidth}
    \centering
    \includegraphics[width=0.9\linewidth]{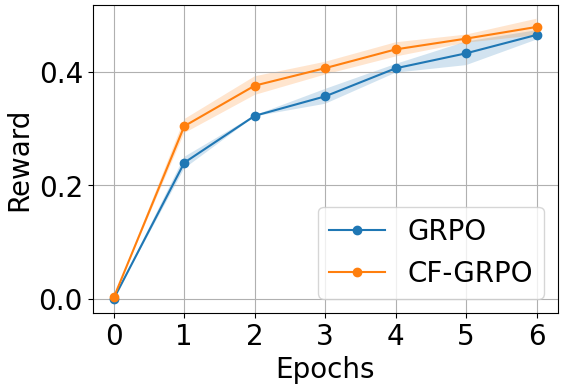}
    \caption{\textbf{CF-RL yields an initial reward boost beyond standard RL.}
CF-GRPO denotes CF-RL with the GRPO algorithm.
Shaded regions denote standard deviation over three runs.}
    \label{fig:cf-grpo}
\end{minipage}
\hfill
    \begin{minipage}{0.49\columnwidth}
        \centering
    \includegraphics[width=0.9\linewidth]{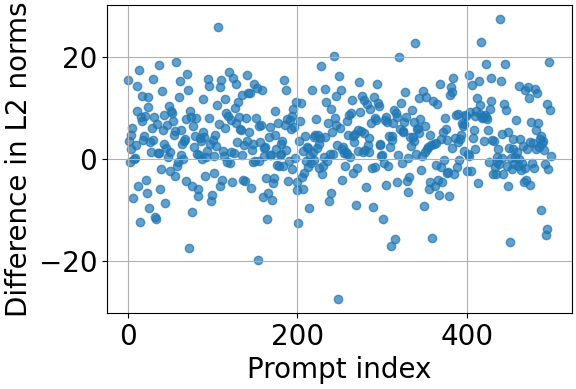}
    \caption{\textbf{CF-RL does not reduce feature changes.}
Each point shows the difference $\|\Delta \phi_{\mathrm{CF\text{-}GRPO}}(x_{i})\| - \|\Delta \phi_{\mathrm{GRPO}}(x_{i})\|$ for prompt $x_{i}$.}
    \label{fig:feature_difference}
    \end{minipage}
\end{figure}

\paragraph{CF-RL does not reduce feature distortion.}
We first examine whether a similar mechanism explains the behavior of CF-RL.
\Cref{fig:feature_difference} plots the difference in feature changes,
$\|\Delta \phi_{\mathrm{CF\text{-}GRPO}}(x_{i})\| - \|\Delta \phi_{\mathrm{GRPO}}(x_{i})\|$,
where $\Delta \phi_{\mathrm{CF\text{-}GRPO}}(x_{i})$ and $\Delta \phi_{\mathrm{GRPO}}(x_{i})$ denote the changes in feature representations from the SFT model for a given prompt $x_{i}$.
The distribution is centered around zero, indicating that CF-RL does not reduce the magnitude of feature changes compared to standard RL.
This suggests that the performance gains of CF-RL cannot be explained by the feature distortion reduction mechanism observed in LP-FT.

\paragraph{CF-RL does not amplify classifier norms.}
Since the only difference between CF-RL and standard RL lies in the classifier initialization at the start of RL, we next examine the properties of the classifier.
We measure the overall norm of the classifier parameters across different training stages.
As shown in \Cref{tab:classifier_norm} (Appendix), the classifier norm remains nearly unchanged across SFT, GRPO, the CF stage, and CF-GRPO.
This behavior contrasts sharply with LP-FT, where classifier norms increase substantially during the probing stage~\citep{tomihari2024understanding}, indicating that CF-RL does not operate by amplifying the scale of the classifier.

\begin{table}[t]
\centering
\caption{Top-5 tokens with the largest classifier updates for each method.}
\label{tab:classifier_tokens}
\begin{tabular}{l l l}
\toprule
\textbf{GRPO} & \textbf{CF-stage} & \textbf{CF-GRPO} \\
\midrule
bringing        & \texttt{<|endoftext|>} & \texttt{<|endoftext|>} \\
Several         & Supplementary           & Appendix \\
it              & Whilst                  & \texttt{\textbackslash x0c} \\
Conduct         & Ibid                    & \texttt{\textbackslash x9d} \\
biomark         & \texttt{\textbackslash n} & Notice \\
\bottomrule
\end{tabular}
\end{table}

\paragraph{Token-level structure of classifier updates.}
We analyze classifier updates at the token level to understand how CF-RL reshapes model predictions.
Each row $W_{v,:} \in \mathbb{R}^D$ of the classifier is used to compute the logit of a single token $v$.
As a result, row-wise updates directly correspond to changes in the model's predicted score for that specific token, independently of other tokens.
\Cref{tab:classifier_tokens} lists the tokens associated with the largest row-wise classifier updates.
The results reveal a qualitative difference between training methods:
standard GRPO primarily amplifies content-bearing tokens, whereas both the CF stage and CF-GRPO exhibit large updates on tokens that encode document structure, formatting, or meta-level information.
These include special tokens, control characters, and markers such as \texttt{<|endoftext|>}, section headers, and footnote-related symbols.
Overall, this indicates that CF-RL reshapes the classifier to emphasize structural and non-semantic cues rather than lexical content.

\paragraph{Summary.}
These findings suggest that CF-RL improves learning not by suppressing feature changes or increasing the overall scale of the classifier.
Instead, during the first stage of CF-RL, the classifier is updated in a way that yields relatively larger changes on rows corresponding to tokens associated with structural information.
Because the classifier appears explicitly in the Gradient component of our analysis, these differences modify the resulting Gradient component, leading to more effective optimization in subsequent RL.


\section{Conclusion}
We analyze the learning dynamics of RL post-training, extending perspectives that have been studied in supervised learning but remain underexplored in RL.
Our formulation provides an explanation for the increase in model confidence based on the limited variability of feature representations.
This analysis highlights the importance of rapidly shaping the classifier, which motivates our proposed method, classifier-first reinforcement learning (CF-RL).
Our experiments validate the theoretical analysis by demonstrating both increased model confidence and accelerated optimization under CF-RL.
Further experimental analysis reveals that the mechanism underlying CF-RL differs from that of LP-FT in supervised fine-tuning.






\bibliographystyle{plainnat}
\bibliography{ref}
\newpage
\appendix
\onecolumn

\renewcommand{\thetable}{S.\arabic{table}}
\renewcommand{\thefigure}{S.\arabic{figure}}
\setcounter{table}{0}
\setcounter{figure}{0}
\renewcommand{\theHtable}{S.\arabic{table}}
\renewcommand{\theHfigure}{S.\arabic{figure}}

\section{Proof}
In the following we use slightly simplified notation:
\[
\pi_\theta(y_l\mid x,y_{<l}) := [\pi_\theta(\cdot\mid x,y_{<l})]_{y_l}.
\]
\subsection{Proof of Proposition~\ref{prop:dynamics}}
\begin{em}
\paragraph{Proposition~\ref{prop:dynamics} (restated).}
The change in the model output can be written as
\begin{align}
\Delta_{t}\log \pi(\cdot \mid x, y_{<m})
    &=  \frac{\eta}{N}\sum_{i=1}^{N}
       \mathbb{E}_{y_{i} \sim \pi_{\theta_{t}}(\cdot \mid x_{i})}
       \Biggl[
           \sum_{l=1}^{|y_{i}|}
           \hat{r}(x_{i},y_{i})\,
           T_{x, y_{<m}}\,\mathcal{K}_{t}(x,y_{<m},x_{i},y_{i,<l})\,d_{y_{i},l}
       \Biggr]
       + O\!\left(\eta^{2}\bigl\|\tfrac{\partial J(\theta_{t})}{\partial \theta}\bigr\|^{2}\right),
\end{align}
Here $\mathcal{K}_{t}(x,y_{<m},x_{i},y_{i,<l})$ is the empirical NTK which is decomposed as 
\begin{align}
     R_{t}(x,y_{<m},x_{i},y_{i,<l}) + G_{t}(x,y_{<m},x_{i},y_{i,<l}),
\end{align}
where $R_{t}(x,y_{<m},x_{i},y_{i,<l})$ is the Representation component
\begin{align}
    \langle\phi_{t}(x,y_{<m}), \phi_{t}(x_{i}, y_{i,<l})\rangle I_{V}
\end{align}
and $G_{t}(x,y_{<m},x_{i},y_{i,<l})$ is the Gradient component 
\begin{align}
    W_{t}\,\frac{\partial \phi_{t}(x, y_{<m})}{\partial\theta^{\phi}}
      \left(\frac{\partial \phi_{t}(x_{i},y_{i,<l})}{\partial\theta^{\phi}}\right)^{\!\top}
      W_{t}^{\top}.
\end{align}
\end{em}

\begin{proof}
We begin by applying a first-order Taylor expansion to the log-policy:
\begin{align}
    \Delta_{t}\log \pi(\cdot \mid x, y_{<m})
    &= \log \pi_{\theta_{t+1}}(\cdot \mid x, y_{<m})
       - \log \pi_{\theta_{t}}(\cdot \mid x, y_{<m}) \\
    &= \frac{\partial \log \pi_{\theta_{t}}(\cdot \mid x, y_{<m})}{\partial \theta}
       (\theta_{t+1} - \theta_{t})
       + O(\|\theta_{t+1}-\theta_{t}\|^{2})\label{eq:taylor}.
\end{align}

\paragraph{Bounding the second-order term.}
Using the update rule $\theta_{t+1}-\theta_{t}=\eta\,\partial J(\theta_{t})/\partial\theta$, we obtain
\begin{align}
    O(\|\theta_{t+1}-\theta_{t}\|^{2})
    &= O\!\left(\|\eta\,\tfrac{\partial J(\theta_{t})}{\partial \theta}\|^{2}\right) \\
    &= O\!\left(\eta^{2}\,\|\tfrac{\partial J(\theta_{t})}{\partial \theta}\|^{2}\right).
\end{align}

\paragraph{Gradient of the objective.}
We expand the reward expectation as
\begin{align}
    \mathbb{E}_{y\sim\pi_{\theta}(\cdot \mid x_{i})}[\hat{r}(x_{i},y)]
    &= \mathbb{E}_{y\sim\pi_{\theta}(\cdot \mid x_{i})}
       \left[r(x_{i},y)
       -\lambda\,\mathrm{KL}\!\left(\pi_{\theta}(\cdot \mid x_{i})\|
            \pi_{\theta_{\mathrm{ref}}}(\cdot \mid x_{i})\right)\right] \\
    &= \mathbb{E}_{y\sim\pi_{\theta}(\cdot \mid x_{i})}
       \left[r(x_{i},y)
       -\lambda\,\frac{\log \pi_{\theta}(y\mid x_{i})}
                       {\log \pi_{\theta_{\mathrm{ref}}}(y\mid x_{i})}\right].
\end{align}
Applying the log-derivative trick, we have
\begin{align}
    \frac{\partial J(\theta)}{\partial \theta}
    &= \frac{1}{N}\sum_{i=1}^{N}
       \frac{\partial}{\partial\theta}\,
       \mathbb{E}_{y\sim\pi_{\theta}(\cdot \mid x_{i})}[\hat{r}(x_{i},y)] \\
    &= \frac{1}{N}\sum_{i=1}^{N}
       \left(
           \mathbb{E}_{y\sim\pi_{\theta}(\cdot \mid x_{i})}
           \!\left[\hat{r}(x_{i},y)\frac{\partial \log \pi_{\theta}(y\mid x_{i})}{\partial \theta}\right]
           -\lambda\,
           \mathbb{E}_{y\sim\pi_{\theta}(\cdot \mid x_{i})}
           \!\left[\frac{\partial \log \pi_{\theta}(y\mid x_{i})}{\partial \theta}\right]
       \right).
\end{align}

Since
\begin{align}
    \mathbb{E}_{y\sim\pi_{\theta}(\cdot \mid x_{i})}
    \left[\frac{\partial \log \pi_{\theta}(y\mid x_{i})}{\partial\theta}\right]
    &= \sum_{y\in\mathcal{Y}}\frac{\partial \pi_{\theta}(y\mid x_{i})}{\partial\theta} \\
    &= \frac{\partial}{\partial\theta}\sum_{y\in\mathcal{Y}}\pi_{\theta}(y\mid x_{i}) \\
    &= \frac{\partial}{\partial\theta}1 \\
    &= 0,
\end{align}
the gradient simplifies to
\begin{align}
    \frac{\partial J(\theta)}{\partial \theta}
    &= \frac{1}{N}\sum_{i=1}^{N}
       \mathbb{E}_{y\sim\pi_{\theta}(\cdot \mid x_{i})}
       \left[\hat{r}(x_{i},y)
       \frac{\partial \log \pi_{\theta}(y\mid x_{i})}{\partial \theta}\right].
\end{align}

\paragraph{NTK computation.}
The model outputs
\begin{align}
    \pi_{\theta}(\cdot \mid x, y_{<l})
    = \mathrm{softmax}(W\phi(x, y_{<l})),
\end{align}
and we denote logits by $f(x,y_{<l})\coloneqq W\phi(x,y_{<l})$.

We compute the NTK matrix:
\begin{align}
\mathcal{K}_{t}(x,y_{<m},x_{i},y_{i,<l})
&\coloneqq
    \frac{\partial f_{t}(x, y_{<m})}{\partial \theta}
    \left(\frac{\partial f_{t}(x_{i}, y_{i,<l})}{\partial \theta}\right)^{\top} \\
    &=  \frac{\partial f_{t}(x, y_{<m})}{\partial \theta^{W}}
        \left(\frac{\partial f_{t}(x_{i}, y_{i,<l})}{\partial \theta^{W}}\right)^{\top}
       + \frac{\partial f_{t}(x, y_{<m})}{\partial \theta^{\phi}}
        \left(\frac{\partial f_{t}(x_{i}, y_{i,<l})}{\partial \theta^{\phi}}\right)^{\top} \\
    &= (\phi_{t}(x, y_{<m})^{\top}\otimes I_{V})(\phi_{t}(x_{i}, y_{i,<l})^{\top}\otimes I_{V})^{\top}
       + W\,\frac{\partial \phi_{t}(x, y_{<m})}{\partial \theta^{\phi}}
         (W\,\frac{\partial \phi_{t}(x_{i}, y_{i,<l})}{\partial \theta^{\phi}})^{\top} \\
    &= \langle\phi_{t}(x, y_{<m}), \phi_{t}(x_{i}, y_{i,<l})\rangle I_{V}
       + W\,\frac{\partial \phi_{t}(x, y_{<m})}{\partial\theta^{\phi}}
      \left(\frac{\partial \phi_{t}(x_{i},y_{i,<l})}{\partial\theta^{\phi}}\right)^{\!\top}
      W^{\top}.
\end{align}
Therefore, the NTK $\mathcal{K}_{t}(x,y_{<m},x_{i},y_{i,<l})$ can be decomposed as
\begin{align}
    \mathcal{K}_{t}(x,y_{<m},x_{i},y_{i,<l}) = R_{t}(x,y_{<m},x_{i},y_{i,<l}) + G_{t}(x,y_{<m},x_{i},y_{i,<l}),
\end{align}
where $R_{t}(x,y_{<m},x_{i},y_{i,<l})$ is the Representation component
\begin{align}
    \langle\phi_{t}(x,y_{<m}), \phi_{t}(x_{i}, y_{i,<l})\rangle I_{V}
\end{align}
and $G_{t}(x,y_{<m},x_{i},y_{i,<l})$ is the Gradient component 
\begin{align}
    W_{t}\,\frac{\partial \phi_{t}(x, y_{<m})}{\partial\theta^{\phi}}
      \left(\frac{\partial \phi_{t}(x_{i},y_{i,<l})}{\partial\theta^{\phi}}\right)^{\!\top}
      W_{t}^{\top}.
\end{align}

\paragraph{First term.}
Combining the computed components, we have
\begin{align}
    &\frac{\partial \log \pi_{\theta_{t}}(\cdot \mid x, y_{<m})}{\partial \theta}
      (\theta_{t+1}-\theta_{t}) \\
    &= \eta\,
       \frac{\partial \log \pi_{\theta_{t}}(\cdot \mid x, y_{<m})}{\partial f(x, y_{<m})}
       \frac{\partial f_{t}(x, y_{<m})}{\partial \theta}
       \left(\frac{\partial J(\theta_{t})}{\partial \theta}\right)^{\!\top} \\
    &= \frac{\eta}{N}\sum_{i=1}^{N}
       \frac{\partial \log \pi_{\theta_{t}}(\cdot \mid x, y_{<m})}{\partial f(x, y_{<m})}
       \frac{\partial f_{t}(x, y_{<m})}{\partial \theta}\,
       \mathbb{E}_{y_{i}\sim\pi_{\theta_{t}}(\cdot \mid x_{i})}
       \!\left[
           \hat{r}(x_{i},y_{i})
           \left(\frac{\partial \log \pi_{\theta_{t}}(y_{i}\mid x_{i})}{\partial \theta}\right)^{\!\top}
       \right] \\
    &= \frac{\eta}{N}\sum_{i=1}^{N}
       \mathbb{E}_{y_{i}\sim\pi_{\theta_{t}}(\cdot \mid x_{i})}
       \left[
           \sum_{l=1}^{|y_{i}|}
           \hat{r}(x_{i},y_{i})
           \frac{\partial \log \pi_{\theta_{t}}(\cdot \mid x, y_{<m})}{\partial f(x, y_{<m})}
           \frac{\partial f_{t}(x, y_{<m})}{\partial \theta}
           \left(
               \frac{\partial \log \pi_{\theta_{t}}(y_{i,l}\mid x_{i},y_{i,<l})}{\partial\theta}
           \right)^{\!\top}
       \right] \\
    &= \frac{\eta}{N}\sum_{i=1}^{N}
       \mathbb{E}_{y_{i}\sim\pi_{\theta_{t}}(\cdot \mid x_{i})}
       \left[
           \sum_{l=1}^{|y_{i}|}
           \hat{r}(x_{i},y_{i})
           \frac{\partial \log \pi_{\theta_{t}}(\cdot \mid x, y_{<m})}{\partial f(x, y_{<m})}
           \frac{\partial f_{t}(x, y_{<m})}{\partial \theta}
           \left(\frac{\partial f_{t}(x_{i},y_{i,<l})}{\partial \theta}\right)^{\top}
           \left(
               \frac{\partial \log \pi_{\theta_{t}}(y_{i,l}\mid x_{i},y_{i,<l})}{\partial f(x_{i},y_{i,<l})}
           \right)^{\!\top}
       \right]
\end{align}

Using the notations
\begin{align}
    \frac{\partial \log \pi_{\theta_{t}}(\cdot \mid x,y_{<m})}{\partial f(x,y_{<m})}
    &= I_{D} - 1\,\pi_{\theta_{t}}(\cdot \mid x,y_{<m})^{\!\top}
      \eqqcolon T_{\pi_{\theta_{t}}(\cdot \mid x,y_{<m})}, \\
    \left(
        \frac{\partial \log \pi_{\theta_{t}}(y_{i,l}\mid x_{i},y_{i,<l})}
             {\partial f(x_{i},y_{i,<l})}
    \right)^{\!\top}
    &= e_{y_{i,l}} - \pi_{\theta_{t}}(\cdot \mid x_{i},y_{i,<l})
      \eqqcolon d_{y_{i},l},
\end{align}
we obtain the form
\begin{align}
    &\frac{\partial \log \pi_{\theta_{t}}(\cdot \mid x,y_{<m})}{\partial\theta}
      (\theta_{t+1}-\theta_{t}) \\
    &= \frac{\eta}{N}\sum_{i=1}^{N}
       \mathbb{E}_{y_{i} \sim \pi_{\theta_{t}}(\cdot \mid x_{i})}
       \bigl[
           \sum_{l=1}^{|y_{i}|}
               \hat{r}(x_{i},y_{i})\;T_{\pi_{\theta_{t}}(\cdot \mid x,y_{<m})}\;\mathcal{K}_{t}(x,y_{<m},x_{i},y_{i,<l})\,d_{y_{i},l}\bigr].
\end{align}

\paragraph{Final expression.}
Substituting the above expressions into Eq.~\eqref{eq:taylor}, we obtain
\begin{align}
    &\Delta_{t}\log \pi(\cdot \mid x, y_{<m}) \\
    &= \frac{\eta}{N}\sum_{i=1}^{N}
       \mathbb{E}_{y_{i} \sim \pi_{\theta_{t}}(\cdot \mid x_{i})}
       \bigl[
           \sum_{l=1}^{|y_{i}|}
               \hat{r}(x_{i},y_{i})\;T_{\pi_{\theta_{t}}(\cdot \mid x,y_{<m})}\;\mathcal{K}_{t}(x,y_{<m},x_{i},y_{i,<l})\,d_{y_{i},l}
       \bigr]
       + O\!\left(\eta^{2}\|\tfrac{\partial J(\theta_{t})}{\partial \theta}\|^{2}\right).
\end{align}

\end{proof}

\subsection{Proof of Proposition~\ref{prop:representation}}
\begin{em}
\paragraph{Proposition~\ref{prop:representation} (restated)}
Assume that the feature inner product satisfies
    \begin{align}
        \langle\phi_{t}(x,y),\phi_{t}(x',y')\rangle &\geq 0
    \end{align}
    for any $(x, y), (x', y') \in \mathcal{X}\times\mathcal{Y}$.
    Then the update induced by the Representation component satisfies
    \begin{align}
        \argmax_{1 \le v \le V} [u^{\mathrm{Rep}}_{t,l}(x,y_{<m},x_{i},y_{i})]_{v} = y_{i,l}.
    \end{align}
\end{em}

\begin{proof}
Recall that
\begin{align}
    u^{\mathrm{Rep}}_{t,l}(x,y_{<m},x_{i},y_{i})
    =
    \langle \phi_{t}(x,y_{<m}), \phi_{t}(x_{i}, y_{i,<l}) \rangle \,
    T_{x, y_{<m}} d_{y_{i},l}.
\end{align}
We first analyze $d_{y_{i},l}$. By definition,
\begin{align}
    [d_{y_{i},l}]_v &= [e_{y_{i,l}} - \pi_{\theta_t}(\cdot \mid x_{i}, y_{i,<l})]_{v} \\
&= 
\begin{cases}
1 - \pi_{\theta_t}(y_{i,l}\mid x_{i}, y_{i,<l}) & \text{if } v = y_{i,l},\\[2pt]
-\pi_{\theta_t}(v \mid x_{i}, y_{i,<l}) & \text{otherwise}.
\end{cases}
\end{align}
Since $\pi_{\theta_t}(v \mid x_{i}, y_{i,<l}) \in [0,1]$ for all $v$, the coordinate $v=y_{i,l}$ attains the strictly largest value. Hence,
\begin{align}
    \argmax_{1\le v \le V} [d_{y_{i},l}]_v = y_{i,l}.
\end{align}

Next we show that $T_{x, y_{<m}}$ preserves the maximizing index of any vector.  
For any $a \in \mathbb{R}^V$,
\begin{align}
    T_{x, y_{<m}} a
    &= (I_V - \mathbf{1}\,\pi_{\theta_t}(\cdot \mid x,y_{<m})^{\!\top})\, a \\
    &= a - \mathbf{1}\,(\pi_{\theta_t}(\cdot \mid x,y_{<m})^{\!\top} a).
\end{align}
The second term is a scalar multiple of $\mathbf{1}$ and thus shifts all coordinates by the same value. Therefore,
\begin{align}
    \argmax_{1\le v\le V} [T_{x, y_{<m}} a]_v
    = \argmax_{1\le v\le V} a_v.
\end{align}

Finally, since
\[
\langle \phi_{t}(x,y_{<m}), \phi_{t}(x_{i}, y_{i,<l}) \rangle \ge 0
\]
by assumption, multiplying a vector by this nonnegative scalar does not change its maximizing index. Consequently, we have
\begin{align}
    \argmax_{1\le v\le V} [u^{\mathrm{Rep}}_{t,l}(x,y_{<m},x_{i},y_{i})]_v &= \argmax_{1\le v\le V} [\langle \phi_{t}(x,y_{<m}), \phi_{t}(x_{i}, y_{i,<l}) \rangle T_{x, y_{<m}} d_{y_{i},l}]_v \\ 
    &= \argmax_{1\le v\le V} [T_{x, y_{<m}} d_{y_{i},l}]_v \\
    &= \argmax_{1\le v\le V} [d_{y_{i},l}]_v \\
    &= y_{i,l}.
\end{align}
\end{proof}

\clearpage

\section{Experimental details}
\subsection{Details of the RLHF experiment}
\paragraph{Implementation.}
Our implementation follows the experimental setup of \citet{razin2025what}\footnote{\url{https://github.com/princeton-pli/what-makes-good-rm}} and is built on PyTorch~\citep{pytorch} and the Hugging Face Transformers library~\citep{wolf-etal-2020-transformers}.
We use the Adam optimizer~\citep{kingma:adam} for all training stages, including SFT, RL, and the CF stage.
The hyperparameters used in our experiments are summarized in \Cref{tab:hyperparams}.

\begin{table}[H]
\centering
\small
\setlength{\tabcolsep}{7pt}
\renewcommand{\arraystretch}{1.15}
\caption{Hyperparameters used in our RLHF experiments.}
\label{tab:hyperparams}
\begin{tabular}{@{} l l l p{0.44\linewidth} @{}}
\toprule
\textbf{Category} & \textbf{Hyperparameter} & \textbf{Value} & \textbf{Description} \\
\midrule

\multirow{2}{*}{\textbf{Generation}} 
 & $\mathrm{temperature}$ & 1 & Temperature for generation. \\
 & $\mathrm{max\_new\_tokens}$ & 512 & Maximum number of generated tokens. \\
 
\midrule

\multirow{3}{*}{\textbf{SFT}}
 & $\mathrm{num\_train\_epochs}$ & 1 & Number of epochs. \\
 & $\mathrm{batch\_size}$ & 32 & Effective batch size (via gradient accumulation). \\
 & $\mathrm{learning\_rate}$ & $1e-6$ & Learning rate. \\

\midrule

\multirow{6}{*}{\textbf{RL}}
 & $\mathrm{num\_train\_epochs}$ & 6 & Number of epochs. \\
 & $\mathrm{kl\_coef}$ & 0.05 & KL penalty coefficient. \\
 & $\mathrm{batch\_size}$ & 16 & Number of trajectories per policy update (via gradient accumulation). \\
 & $\mathrm{num\_mini\_batches}$ & 2 & Number of minibatches per update. \\
 & $\mathrm{learning\_rate}$ & $1e-7$ & Learning rate. \\
 & $\mathrm{k}$ & 8 (GRPO) / 2 (RLOO) & Number of rollouts per prompt. \\

\midrule

\multirow{2}{*}{\textbf{CF-stage}}
& $\mathrm{num\_train\_epochs}$ & 1 & Number of epochs. \\
 & $\mathrm{learning\_rate}$ & $1e-7$ & Learning rate. \\
 
\bottomrule
\end{tabular}
\end{table}

\paragraph{Dataset.}
For SFT, we used the \textrm{sft} split of AlpacaFarm\footnote{\url{https://huggingface.co/datasets/tatsu-lab/alpaca_farm}}.
For RL, we used the binarized version of UltraFeedback\footnote{\url{https://huggingface.co/datasets/HuggingFaceH4/ultrafeedback_binarized}}, filtering out samples in which the prompt or either output exceeded $512$ tokens according to the Pythia tokenizer, relabeling output preferences using the reward model, and using $20\%$ of the samples in our experiments.
We used the default or original chat template, as illustrated in \Cref{fig:chat-template-example}.

\begin{figure}[H]
  \centering
  \begin{minipage}{0.95\linewidth}
    \begin{lstlisting}[style=chattemplate]
        <|user|>Can consistent exercise and physical activity improve the quality of sleep and reduce insomnia symptoms?<|endoftext|><|assistant|>Yes. In a study it was found that regular exercise and physical activity could improve sleep quality by increasing oxygen flow, heart rate, and blood volume. Regular exercise also helps to reduce fatigue, which in turn leads to improved sleep quality. Furthermore, consistent exercise was found to reduce anxiety and improve resilience, which in turn can help to reduce the symptoms of insomnia.<|endoftext|>
    \end{lstlisting}
    \subcaption{Default chat template}
\end{minipage}
  \begin{minipage}{0.95\linewidth}
  \begin{lstlisting}[style=chattemplate]
<|start_header_id|>user<|end_header_id|>

Can consistent exercise and physical activity improve the quality of sleep and reduce insomnia symptoms?<|eot_id|><|start_header_id|>assistant<|end_header_id|>

Yes. In a study it was found that regular exercise and physical activity could improve sleep quality by increasing oxygen flow, heart rate, and blood volume. Regular exercise also helps to reduce fatigue, which in turn leads to improved sleep quality. Furthermore, consistent exercise was found to reduce anxiety and improve resilience, which in turn can help to reduce the symptoms of insomnia.<|eot_id|>
  \end{lstlisting}
  \subcaption{Chat template used in ArmoRM}
  \end{minipage}
  \caption{Examples of chat-formatted text.}
  \label{fig:chat-template-example}
\end{figure}

\paragraph{Reward normalization.}
Following prior studies~\citep{razin2025what,gao2023scaling}, we normalized rewards to a common scale.
We sampled $500$ prompts from the policy gradient training set and generated $10$ outputs per prompt using the initial policy, yielding $5000$ outputs in total.
We computed the mean and standard deviation over these outputs and used them to shift and normalize rewards during training and evaluation.

\subsection{Hardware}
All experiments were conducted on an NVIDIA Grace–Hopper system, consisting of an NVIDIA Grace CPU (120 GB memory) and an NVIDIA Hopper H100 GPU (96 GB memory).

\subsection{Details of figures and tables}
\label{app:detail_fig}
\paragraph{Table~\ref{tab:feature}.}
We used the model after SFT. All metrics were computed on $500$ samples from the UltraFeedback dataset. Feature vectors were evaluated only on the prompts, without any generated continuations; specifically, we used $\phi(x_i)$ for each prompt $x_i$.

\paragraph{Figure~\ref{fig:first_token_entropy}.}
We computed the entropy of the first generated token, i.e., $\pi_{\theta_t}(x_i)$, for each prompt $x_i$ using $500$ samples from the UltraFeedback dataset.

\section{Additional experimental results}

\subsection{Results with RLOO}
\begin{figure}[H]
    \centering
    \includegraphics[width=0.9\linewidth]{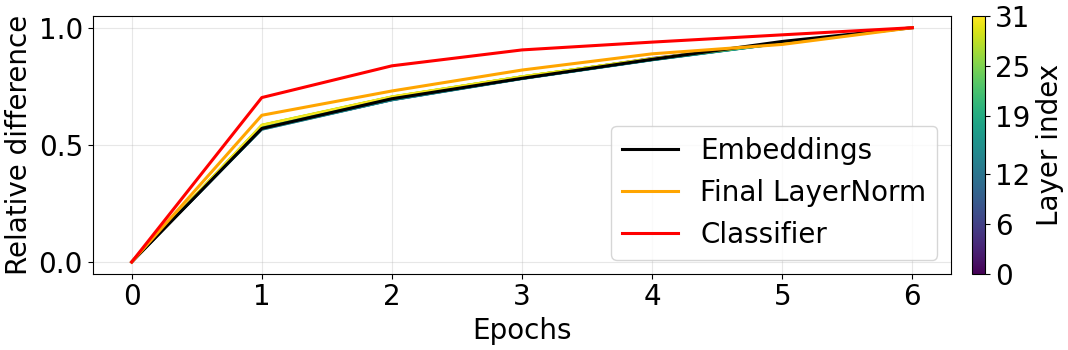}
    \caption{
\textbf{The classifier learns faster than the other parameters.}
This corresponds to \Cref{fig:weight_change}, with RLOO used instead of GRPO.
}
\label{fig:weight_change_rloo}
\end{figure}

\begin{figure}[H]
    \centering
    \includegraphics[width=0.33\linewidth]{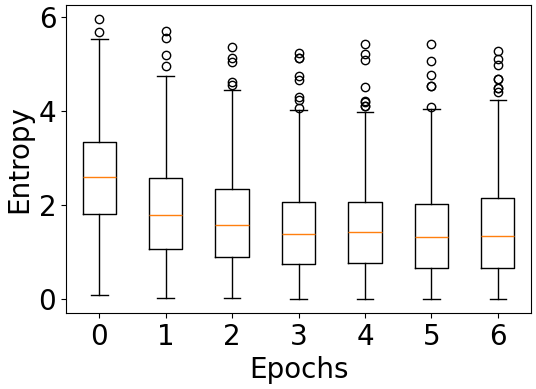}
    \caption{Entropy reduction during RL. Epoch~0 corresponds to the SFT model. This corresponds to \Cref{fig:first_token_entropy}, with RLOO used instead of GRPO.}
\label{fig:first_token_entropy_rloo}
\end{figure}

\begin{figure}[H]
    \centering
    \begin{minipage}{0.48\linewidth}
        \includegraphics[width=0.9\linewidth]{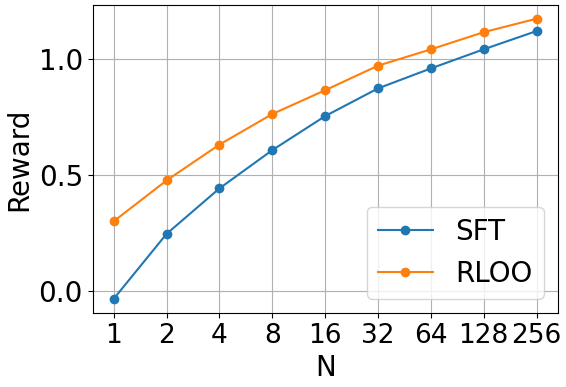}
        \subcaption{Best-of-$N$ rewards for SFT and RLOO.}
        \label{fig:bestofn_rloo}
    \end{minipage}
    \hfill
    \begin{minipage}{0.48\linewidth}
        \includegraphics[width=0.9\linewidth]{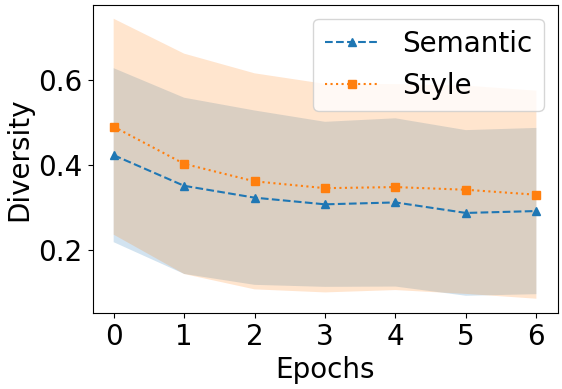}
        \subcaption{Semantic and style diversity over RL epochs. Higher values indicate greater diversity. Shaded regions denote standard deviation.}
        \label{fig:diversity_rloo}
    \end{minipage}
    \caption{\textbf{RL improves reward by concentrating probability mass, at the cost of reduced output diversity.} This corresponds to \Cref{fig:BoN-diversity}, with RLOO used instead of GRPO.}
    \label{fig:BoN-diversity_rloo}
\end{figure}

\begin{figure}[H]
    \centering
    \includegraphics[width=0.50\linewidth]{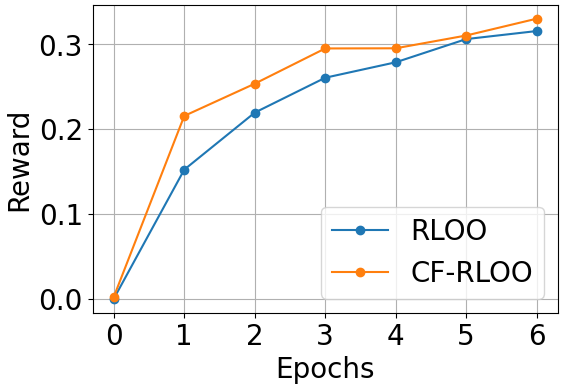}
    \caption{\textbf{CF-RL yields an initial reward boost beyond standard RL.}
CF-RLOO denotes CF-RL with the RLOO algorithm.
Shaded regions denote standard deviation over three runs. This figure corresponds to \Cref{fig:cf-grpo}, with RLOO used instead of GRPO.}
    \label{fig:cf-rloo}
\end{figure}

\begin{figure}[H]
    \centering
    \includegraphics[width=0.50\linewidth]{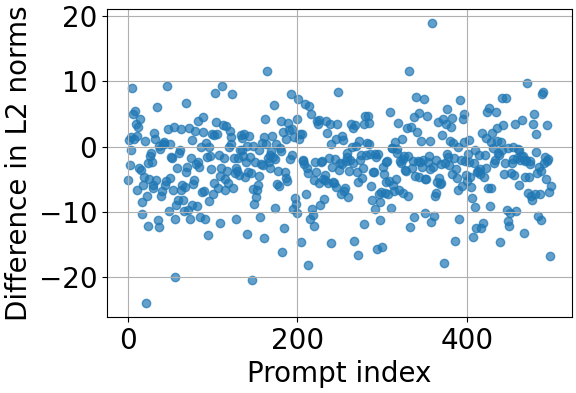}
    \caption{\textbf{CF-RL does not reduce feature changes.}
Each point shows the difference $\|\Delta \phi_{\mathrm{CF\text{-}RLOO}}(x_{i})\| - \|\Delta \phi_{\mathrm{RLOO}}(x_{i})\|$ for prompt $x_{i}$. This corresponds to \Cref{fig:cf-grpo}, with RLOO used instead of GRPO.}
    \label{fig:feature_difference_rloo}
\end{figure}

We show additional experiments conducted with RLOO~\citep{kool2019buy,ahmadian2024back} instead of GRPO.
\begin{itemize}
    \item \Cref{fig:weight_change_rloo} corresponds to \Cref{fig:weight_change}
    (parameter-wise weight change during RL).
    \item \Cref{fig:first_token_entropy_rloo} corresponds to \Cref{fig:first_token_entropy}
    (first-token entropy across training epochs).
    \item \Cref{fig:BoN-diversity_rloo} corresponds to \Cref{fig:BoN-diversity}
    (Best-of-N and diversity).
    \item \Cref{fig:cf-rloo} corresponds to \Cref{fig:cf-grpo}
    (comparison between standard RL and CF-RL).
    \item \Cref{fig:feature_difference_rloo} corresponds to \Cref{fig:feature_difference}
    (difference in feature changes between CF-RL and standard RL).
    \item \Cref{tab:classifier_tokens_rloo} corresponds to \Cref{tab:classifier_tokens} (Top tokens with the largest classifier updates).
\end{itemize}

\subsection{Classifier norm}
\Cref{tab:classifier_norm} shows the norm of the classifier after SFT, GRPO, the CF stage, and CF-GRPO.
The classifier norm remains nearly unchanged across all training.

\begin{table}[H]
\centering
\caption{Top-5 tokens with the largest classifier updates for each method. As with GRPO, standard RLOO mainly amplifies content-bearing tokens, whereas CF-RLOO yields large updates on tokens that encode document structure, formatting, or meta-level information (e.g., special tokens and control characters).
}
\label{tab:classifier_tokens_rloo}
\begin{tabular}{l l}
\toprule
\textbf{RLOO} & \textbf{CF-RLOO} \\
\midrule
glanced        & \texttt{<|endoftext|>} \\
effect         & \texttt{\textbackslash n\textbackslash x0c} \\
No             & Supplementary \\
NOTES          & \texttt{[…]} \\
Tags           & \texttt{<U+FFFD>} \\
\bottomrule
\end{tabular}
\end{table}

\begin{table}[H]
\centering
\caption{Mean and standard deviation of classifier update norms.}
\label{tab:classifier_norm}
\begin{tabular}{lcc}
\toprule
Method & Mean & Std \\
\midrule
SFT        & 0.9892 & 0.1175 \\
GRPO       & 0.9892 & 0.1175 \\
CF-stage   & 0.9893 & 0.1175 \\
CF-GRPO    & 0.9893 & 0.1175 \\
\bottomrule
\end{tabular}
\end{table}

\end{document}